\documentclass[twoside,11pt]{article}
\DeclareUnicodeCharacter{2717}{\ding{55}}
\DeclareUnicodeCharacter{2713}{\ding{51}}
\usepackage{jmlr2e}
\usepackage{listings}
\usepackage{subfigure}
\usepackage{booktabs}
\usepackage{minted} 
\usepackage{multirow}
\usepackage{pifont}
\newcommand{\eg}{\emph{e.g.}}
\newcommand{\ie}{\emph{i.e.}}
\newcommand{\name}{{\sc C3Box }}
\newcommand{\mame}{{\sc C3Box}}

\firstpageno{1}

\begin{document}

\title{C3Box: A CLIP-based Class-Incremental Learning Toolbox}

\author{\name Hao Sun \email sunhao@lamda.nju.edu.cn \\
	\name Da-Wei Zhou \email zhoudw@lamda.nju.edu.cn \\
       \addr School of Artificial Intelligence, Nanjing University, China\\
       National Key Laboratory for Novel Software Technology, Nanjing University, 210023, China}
\maketitle

\begin{abstract}
Traditional machine learning systems are typically designed for static data distributions, which suffer from catastrophic forgetting when learning from evolving data streams. Class-Incremental Learning (CIL) addresses this challenge by enabling learning systems to continuously learn new classes while preserving prior knowledge. With the rise of pre-trained models (PTMs) such as CLIP, leveraging their strong generalization and semantic alignment capabilities has become a promising direction in CIL. However, existing CLIP-based CIL methods are often scattered across disparate codebases, rely on inconsistent configurations, hindering fair comparisons, reproducibility, and practical adoption. Therefore, we propose \name(\underline{\textbf{C}}LIP-based \underline{\textbf{C}}lass-in\underline{\textbf{C}}remental
learning tool\underline{\textbf{BOX}}), a modular and comprehensive Python toolbox. \name integrates representative traditional CIL methods, ViT-based CIL methods, and state-of-the-art CLIP-based CIL methods into a unified CLIP-based framework. By inheriting the streamlined design of PyCIL, \name provides a JSON-based configuration and standardized execution pipeline. This design enables reproducible experimentation with low engineering overhead and makes \name a reliable benchmark platform for continual learning research.  Designed to be user-friendly, \name relies only on widely used open-source libraries and supports major operating systems.  
The code is available at~\url{https://github.com/LAMDA-CL/C3Box}.

\end{abstract}


\section{Introduction}
In recent years, the rapid advancement of deep learning~\citep{liu2015deep} has demonstrated immense potential across numerous domains. However, real-world scenarios are inherently dynamic, with data typically emerging as continuous and evolving data streams~\citep{gomes2017survey}. Traditional deep learning methods are typically optimized for static data distributions and struggle to integrate new classes over time. When learning from dynamic data streams, these methods often overwrite previous knowledge to accommodate new classes, resulting in catastrophic forgetting~\citep{french1999catastrophic}. Class-Incremental Learning (CIL)~\citep{rebuffi2017icarl} has been proposed to address this issue~\citep{zhou2024class}. Driven by the remarkable success of Pre-trained Models (PTMs),  such as Vision Transformers~\citep{dosovitskiy2020image} and CLIP~\citep{radford2021learning}, the focus of CIL research has undergone a significant shift. The field is moving away from traditional training from scratch with randomly initialized weights toward leveraging the inherent generalization capabilities of PTMs~\citep{zhou2024continual}.

In particular, CLIP~\citep{radford2021learning} has emerged as a pioneering pre-trained vision-language model~\citep{wang2024survey} that provides a powerful starting point for continual learning by aligning visual concepts with natural language in a shared embedding space. Unlike traditional vision-only models, CLIP leverages rich textual semantics to guide the learning process, offering a more robust representation that effectively mitigates catastrophic forgetting while adapting to dynamic real-world scenarios~\citep{hu2025hierarchical,wen2025hierarchical}. Despite these advantages, current CLIP-based CIL research still faces substantial practical challenges. Firstly, the implementation of various CLIP-based CIL methods is highly fragmented, with researchers often relying on disparate codebases and inconsistent experimental protocols. This lack of a unified framework makes it difficult to reproduce results and build upon previous work. Secondly, the absence of a standardized experimental protocol leads to significant evaluation inconsistency, where the use of varying data splits and evaluation metrics prevents fair comparisons. Thirdly, integrating CLIP-based methods into diverse incremental learning scenarios remains cumbersome, as different adaptation strategies often require bespoke modules and interface engineering, which imposes a significant engineering burden on researchers.
\begin{figure}[t] \label{figure:incremental}
	\begin{center}
	\includegraphics[width=1\columnwidth]{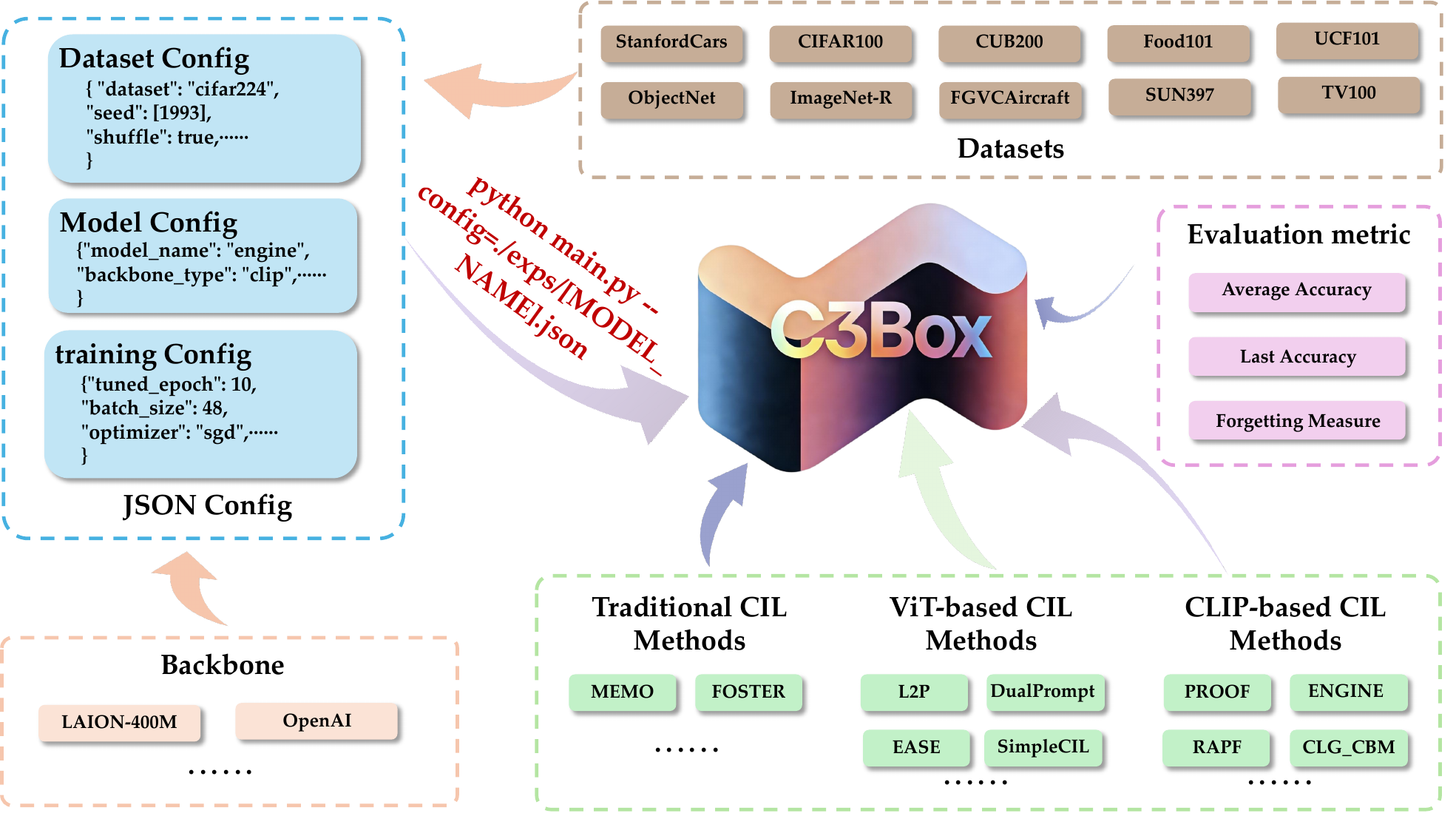}
	\end{center}
	\vspace{-5mm}
	\caption{ Overview of \name and its main functionalities and modules.
	} 	\vspace{-5mm}
\end{figure}

To bridge these gaps and provide a standardized platform for the machine learning community, we present \name (\underline{\textbf{C}}LIP-based \underline{\textbf{C}}lass-in\underline{\textbf{C}}remental
learning tool\underline{\textbf{BOX}}), a modular and comprehensive toolbox specifically tailored. 
As shown in Figure~\ref{figure:incremental}, \name provides extensive algorithm coverage by integrating state-of-the-art CLIP-based CIL methods alongside representative traditional and ViT-based CIL approaches, as well as fundamental baselines, within a unified framework that enables systematic and fair baseline comparisons. Moreover, building on the streamlined architecture of PyCIL~\citep{zhou2023pycil} and PILOT~\citep{sun2025pilot}, \name provides a unified configuration and execution pipeline, allowing users to define datasets, backbones, and training settings in a single JSON file. This design offers a standardized interface that allows for seamless integration of new methods and automated logging, ensuring high reproducibility and minimal engineering overhead. As a user-friendly toolbox, \name adopts consistent unified interfaces across methods and relies only on widely used open-source libraries to ensure easy adoption and broad compatibility across major operating systems, including Linux, macOS, and Windows.

\section{Toolbox Usage}
{\noindent\bf Dependencies:} Building upon the established architectures of PyCIL~\citep{zhou2023pycil} and PILOT~\citep{sun2025pilot}, \name is built on a robust, modular software stack. It relies solely on a suite of widely adopted open-source libraries, such as NumPy~\citep{harris2020array} and SciPy~\citep{virtanen2020scipy} for fundamental numerical operations and optimization, while the core neural architectures are implemented using the PyTorch \citep{paszke2019pytorch} framework. We also utilize the OpenCLIP library \citep{cherti2023reproducible} to provide a standardized interface for loading diverse pre-trained model weights.
The tqdm~\citep{da2019tqdm} library provides real-time progress monitoring.

\noindent\textbf{Supported datasets:} To thoroughly support the evaluation of various algorithms, we follow~\citep{zhou2025learning} and select ten benchmark datasets with significant domain gaps from CLIP’s pre-training dataset. The specific evaluation benchmarks include: CIFAR100 \citep{Krizhevsky2009LearningML}, CUB200 \citep{wah2011caltech}, ObjectNet \citep{barbu2019objectnet}, ImageNet-R \citep{hendrycks2021many}, FGVCAircraft \citep{maji2013fine}, StanfordCars \citep{krause20133d}, Food101 \citep{bossard2014food}, SUN397 \citep{xiao2010sun}, UCF101 \citep{soomro2012ucf101} and TV100~\citep{zhou2024tv100}. Following the benchmark protocols in CIL~\citep{zhou2025learning}, we evaluate on 100 classes for CIFAR100, Aircraft, Cars, Food, UCF and TV100; 200 classes for CUB200, ObjectNet, and ImageNet-R; and 300 classes for SUN to maintain consistency across incremental stages.
\\\textbf{Dataset split:} Following the evaluation protocols in CIL~\citep{zhou2024class}, we utilize `B-$m$ Inc-$n$' to split the classes. Specifically, $m$ denotes the number of classes in the initial stage, while $n$ represents the number of classes in each subsequent incremental stage. 
\\\textbf{Implemented Methods:} 
\name implements a total of 17 representative CIL methods, covering traditional CIL methods, \ie, FOSTER~\citep{wang2022foster}, MEMO~\citep{zhou2022model}, ViT-based CIL methods, \ie, L2P~\citep{wang2022learning}, DualPrompt~\citep{wang2022dualprompt}, CODA-Prompt~\citep{smith2023coda}, EASE~\citep{zhou2024expandable}, SimpleCIL~\citep{zhou2025revisiting}, APER (with Adapter/Finetune/SSF/VPT variants)~\citep{zhou2025revisiting}, TUNA~\citep{wang2025integrating}, and state-of-the-art CLIP-based methods, including RAPF~\citep{huang2024class}, 
CLG-CBM~\citep{yu2025language}, MG-CLIP~\citep{huang2025mind}, PROOF~\citep{zhou2025learning}, ENGINE~\citep{zhou2025external}, and BOFA~\citep{li2025bofa}. In addition, \name provides common baselines such as Finetune and ZS-CLIP~\citep{radford2021learning}. Notably, all methods have been adapted into a unified CLIP-based framework. Implementation details are provided in Appendix~\ref{methods}.

{\noindent\bf Evaluation metric:} Following the evaluation protocols in CIL~\citep{rebuffi2017icarl,zhou2025learning}, we denote $\mathcal{A}_b$ as the model's accuracy after the $b$-th incremental stage. In \mame, we employ main metrics to evaluate this implemented method: \textbf{Last Accuracy}  $\mathcal{A}_B$, which reflects performance after the last task, and \textbf{Average Accuracy} $\bar{\mathcal{A}}=\frac{1}{B}\sum_{b=1}^{B}\mathcal{A}_b$, which represents the mean accuracy across all incremental stages. In addition, following~\citep{chaudhry2018riemannian}, we define \textbf{Forgetting Measure
}  as the average drop from the best-achieved accuracy
of each task to its last accuracy:
\begin{equation}
      F_{B} = \frac{1}{B-1}\sum_{b=1}^{B-1}\max_{l \in \{b, \dots, B-1\}} (\mathcal{A}_{l,b}- \mathcal{A}_{B,b}).
\end{equation}

\begin{table}[t]
\vspace{-5mm}
    \caption{Average and last performance of different methods on CIFAR100 B0 Inc10 and Aircraft B0 Inc10. `-' indicates the original paper didn't report the performance.}
    \label{tab:benchmark}
    \centering
    \tabcolsep 5pt 
    \resizebox{\textwidth}{!}{
        \begin{tabular}{clccccccccc}
            \toprule
            \multirow{3}{*}{} & \multirow{3}{*}{Method} & \multirow{3}{*}{Exemplars} 
            & \multicolumn{4}{c}{CIFAR100} 
            & \multicolumn{4}{c}{Aircraft} 
            \\
            & & & \multicolumn{2}{c}{Reproduced} 
            & \multicolumn{2}{c}{Reported}
            & \multicolumn{2}{c}{Reproduced} 
            & \multicolumn{2}{c}{Reported} 
            \\
            & & & $\bar{\mathcal{A}}$ & $\mathcal{A}_B$ 
            & $\bar{\mathcal{A}}$ & $\mathcal{A}_B$ 
            & $\bar{\mathcal{A}}$ & $\mathcal{A}_B$ 
            & $\bar{\mathcal{A}}$ & $\mathcal{A}_B$ \\
            \midrule
            \multirow{2}{*}{Baselines} 
            & Finetune & ✗ & 21.33  & 9.24 & - & -& 6.22 & 3.42 & - & -  \\
            & ZS-CLIP~\citep{radford2021learning} & ✗ & 81.81  & 71.38 & - & -& 26.61 & 17.16 & - & - \\
            \midrule
            
            \multirow{2}{*}{Traditional} 
            & FOSTER~\citep{wang2022foster}  & ✓ & 86.56 & 80.32 & - & -& 52.96 & 39.87 & -  & - \\
            & MEMO~\citep{zhou2022model} & ✓ & 85.05 & 73.68 & - & -& 42.24  & 25.41 & -  & - \\
            \midrule
            
            \multirow{11}{*}{ViT-based} 
            & L2P~\citep{wang2022learning} & ✓ & 86.53  & 77.41 & - & - & 55.06  &   44.88 & - & - \\
            & DualPrompt~\citep{wang2022dualprompt} & ✓ & 82.67  & 74.86 & - & - &  60.10 &  55.30 & - & - \\
            & CODA-Prompt~\citep{smith2023coda} & ✓ & 86.58  & 77.91 & - & - & 56.57   &  55.81 & - & - \\
            & EASE~\citep{zhou2024expandable} & ✗ & 81.04 & 71.51 & - & - & 57.17 & 45.27 & - & - \\
            & SimpleCIL~\citep{zhou2025revisiting} & ✗ & 84.15 & 76.63 & - & -& 59.06 & 47.94 & -  & - \\
            
             & APER + Finetune~\citep{zhou2025revisiting} & ✗ & 83.74  & 75.37 & - & -  & 56.57  & 44.79 & - & - \\
            & APER + Adapter~\citep{zhou2025revisiting} & ✗ & 84.91  & 76.67 & - & -  & 57.68  & 46.71 & - & - \\
            & APER + SSF~\citep{zhou2025revisiting} & ✗ & 78.60  & 68.44 & - & -  & 58.69  & 47.61 & - & - \\
            & APER + VPT-D~\citep{zhou2025revisiting} & ✗ &  78.27 & 67.87 & - & -  & 63.94  & 51.70 & - & - \\
            & APER + VPT-S~\citep{zhou2025revisiting} & ✗ & 83.76  & 75.66 & - & -  & 57.94  & 47.19 & - & - \\
            & TUNA~\citep{wang2025integrating} & ✗ & 86.39 & 76.89 & - & - & 61.27 & 44.88 & - & - \\
            \midrule
            
            \multirow{6}{*}{CLIP-based} 
           
            & RAPF~\citep{huang2024class} & ✗ & 87.52 & 80.88 & 86.87  & 79.26 & 43.88 & 23.58 & - & - \\
            & CLG-CBM~\citep{yu2025language} & ✗ & 86.58 & 80.15 & 84.49 & 76.82 & 66.05 & 55.93 & - & - \\
            & MG-CLIP~\citep{huang2025mind} & ✗ & 88.69 & 80.69 & 87.00  & 80.57 & 49.96 & 32.73 & - & - \\
            & PROOF~\citep{zhou2025learning} & ✓ & 86.77  & 78.58 & 86.70  & 79.05   & 64.03 & 56.17  & 64.61   & 55.81 \\
            & ENGINE~\citep{zhou2025external} & ✗ & 86.78 & 79.32 & 86.92  & 79.22  & 69.69 & 57.76 & 69.69 & 58.69 \\
            & BOFA~\citep{li2025bofa} & ✗ & 86.07  & 79.18 & 86.50  & 79.34 & 70.91 & 60.43 & 69.94  & 59.67 \\
            \bottomrule
        \end{tabular}
    }
    \vspace{-3mm}
\end{table}

{\noindent\bf Basic Usage:} \name features a highly parameterized and unified management framework. All experimental parameters, ranging from \textbf{dataset specifications} and \textbf{model architectures} to \textbf{training protocols}, are encapsulated in a single, human-readable JSON file. This centralized modular design eliminates the need for modifying underlying code; instead, users can simply adjust the global parameters or method-specific hyperparameters within the JSON file, and then run a standardized command:
\begin{center}
 \vspace{-1mm}
    \texttt{python main.py --config=./exps/[MODEL\_NAME].json}
    \vspace{-1mm}
\end{center}
where \texttt{[MODEL\_NAME]} corresponds to one of the implemented methods in \mame, \eg, finetune, l2p, dual, engine, proof, clg\_cbm, and bofa. The primary global parameters within this framework include:
\begin{itemize}
\item {\bf backbone-type:} The pre-trained backbone weights used for model initialization. Our framework supports two commonly used pre-trained CLIP weight options, including LAION-400M~\citep{ilharco_gabriel_2021_5143773} and OpenAI~\citep{radford2021learning}, for the CLIP with ViT-B/16 backbone.
\item {\bf init-cls:} The number of classes in the initial stage. Our framework provides the flexibility to define various class numbers for the initial stage.
\item {\bf increment:} The number of classes added in each incremental stage $i$, where $i>1$. By default, our framework assumes that the number of classes remains constant across all subsequent incremental stages.
	\item {\bf memory\_per\_class:} The fixed number of exemplars is stored for each former class. For replay-based methods in \mame, \eg, PROOF, MEMO, and FOSTER, following the setting in CIL~\citep{zhou2025learning}, we use
the herding~\citep{welling2009herding} algorithm to select 20 exemplars per class for rehearsal.
 \item {\bf seed:} The random seed adopted for shuffling the class order. Following~\citep{rebuffi2017icarl}, we randomly shuffle the
class order using a random seed of 1993 by default.
\end{itemize}

In addition to the above settings, global parameters, \eg, tuned\_epoch, batch\_size, optimization epoch, learning rate, weight\_decay, init\_lr, optimizer, and method-specific
hyperparameters, can be adjusted in the corresponding JSON file.

\section{Preliminary Experiments}
\begin{figure}[t] 
	\begin{center}
	\includegraphics[width=1\columnwidth, height=0.4\textheight, keepaspectratio]{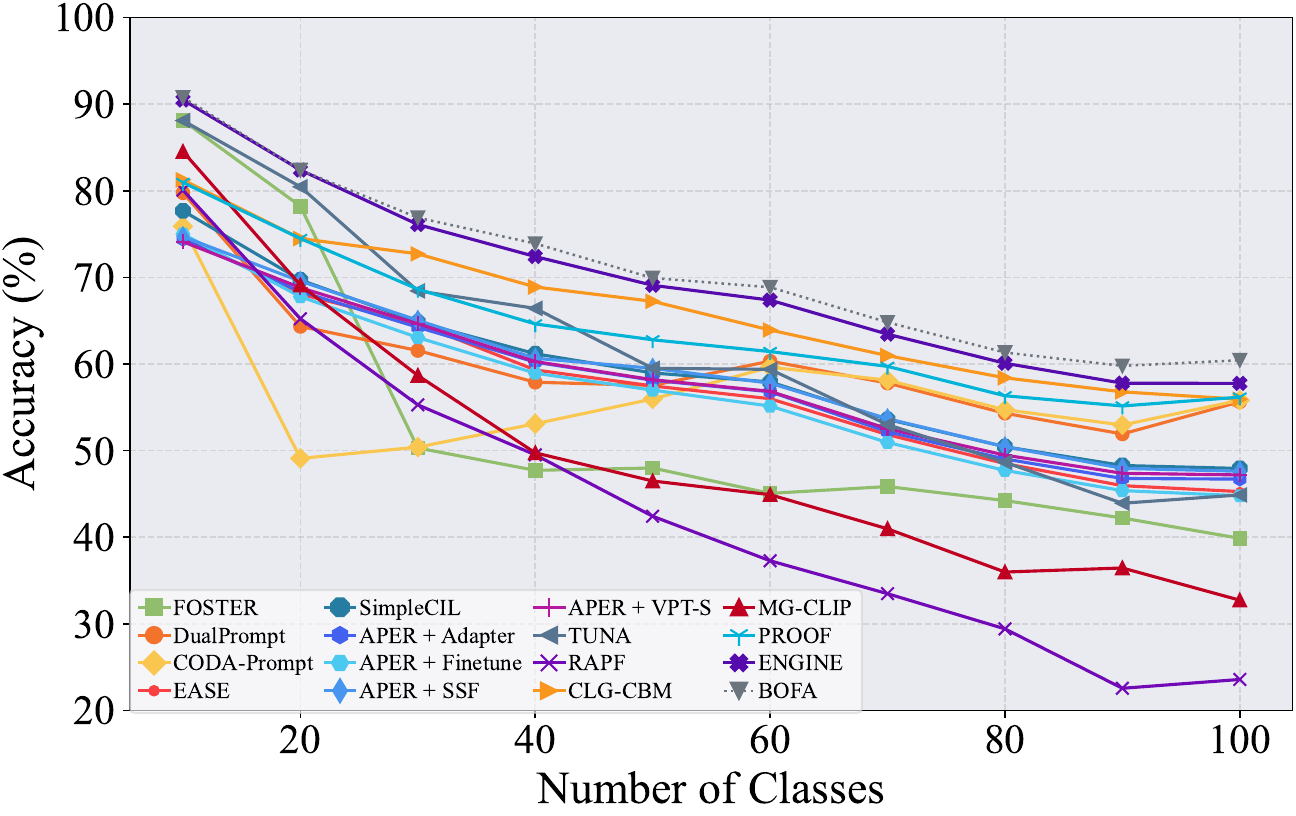}
    \end{center}
	\vspace{-5mm}
	\caption{ Reproduced incremental performance of different methods on Aircraft B0 Inc10.
	} 	\vspace{-5mm}
 \label{fig:c}
\end{figure}
As a preliminary study for the machine learning community, we utilize a single NVIDIA 4090 GPU to evaluate the implemented methods on CIFAR100 B0
Inc10 and Aircraft B0
Inc10, using the LAION-400M pre-trained CLIP model. The results are reported in Table~\ref{tab:benchmark} and Figure~\ref{fig:c}. As the table shows, CLIP-based methods mostly outperform traditional CIL methods, indicating that leveraging CLIP’s strong generalization and semantic alignment helps mitigate catastrophic forgetting in incremental scenarios. 

\section{Conclusion}

We present \mame, a modular toolbox designed to standardize and simplify CLIP-based CIL. By integrating representative methods into a unified framework, \name eliminates implementation fragmentation and ensures fair comparisons. Its cross-platform compatibility and reliance on open source libraries make it an accessible and reliable benchmark for the community. In the future, we will continuously update \name by integrating emerging algorithms and diverse benchmarks to support an even broader range of research.

\bibliography{scis}
\appendix
\section{Implemented Class-Incremental Learning Methods} \label{methods}
We briefly introduce the implemented methods in \mame, including traditional CIL methods, ViT-based methods, and CLIP-based methods. All methods have been adapted into a unified CLIP-based framework. They are listed as:
\begin{itemize}
\item{\bf Finetune:} The baseline method which uses a pre-trained CLIP as model initialization and simply finetunes CLIP on each task, suffering from severe catastrophic forgetting.
\item {\bf ZS-CLIP~\citep{radford2021learning}:} The baseline method which freezes the pre-trained CLIP and predicts class probabilities by computing the cosine similarity, serving as a performance benchmark for the pre-trained CLIP on downstream tasks.
\item {\bf SimpleCIL~\citep{zhou2025revisiting}:} A simple baseline that only utilizes the visual branch of CLIP. It freezes the image encoder and directly extracts prototypes for each new class, using a cosine classifier for prediction.
	\item {\bf L2P~\citep{wang2022learning}:} This method only utilizes the visual branch of CLIP, using visual prompt tuning~\citep{jia2022visual} on the frozen image encoder. It leverages a key-value-based prompt pool to generate instance-specific prompts for task adaptation.
    \item {\bf DualPrompt~\citep{wang2022dualprompt}:} An extension of L2P that only utilizes the visual branch of CLIP. It decouples prompts into general and expert prompts while keeping the other settings the same as L2P.
    \item {\bf CODA-Prompt~\citep{smith2023coda}:} An extension of L2P that only utilizes the
visual branch of CLIP. It replaces key-value prompt selection with a decomposed attention mechanism, using attention weights to guide the recombination of prompts.
     \item {\bf FOSTER~\citep{wang2022foster}:} This method only utilizes the visual branch of CLIP to expand features while maintaining a single backbone. It employs knowledge distillation~\citep{hinton2015distilling} to compress expanded models, effectively alleviating memory costs.
     \item {\bf MEMO~\citep{zhou2022model}:} This method only utilizes the visual branch of CLIP. It decouples the network into specialized and generalized layers. In the implementation, we decouple the vision transformer at the last
transformer block to reduce memory costs during expansion.
    \item{\bf EASE~\citep{zhou2024expandable}:} This method only utilizes the visual branch of CLIP to create task-specific subspaces via lightweight adapters. It employs a semantic-guided prototype complement strategy to synthesize old class features without original instances.
\item{\bf TUNA~\citep{wang2025integrating}:} This method only utilizes the visual branch of CLIP by integrating task-specific and universal adapters. It employs an entropy-based selection mechanism to harness both specialized and shared knowledge during inference.
\item{\bf APER~\citep{zhou2025revisiting}:} This method only utilizes the visual branch of CLIP by aggregating the pre-trained model with an adapted version during the initial stage. It incorporates various Parameter-Efficient Fine-Tuning (PEFT) techniques, including Adapter, Finetune, layer-wise rescale (SSF), and VPT (deep/shallow), to unify generalizability and task-specific adaptivity.
     \item {\bf RAPF~\citep{huang2024class}:} This method combines adaptive representation adjustment based on semantic distances with decomposed parameter fusion on shared orthogonal bases to encode new knowledge into CLIP.
 \item {\bf CLG-CBM~\citep{yu2025language}:} This method utilizes a concept bottleneck layer to align semantics with the CLIP model, enabling the learning of human-understandable and cross-task generalizable concepts.
    \item {\bf MG-CLIP~\citep{huang2025mind}:} This method identifies CLIP’s modality gap to reflect pre-trained knowledge preservation. It introduces modality gap preservation to mitigate forgetting and modality gap compensation to enhance adaptation to new tasks.
    \item {\bf PROOF~\citep{zhou2025learning}:} This method introduces expandable projections and a cross-modal fusion module into CLIP. It aligns visual prototypes with textual features, capturing task-specific semantics while freezing former parameters.
   \item {\bf ENGINE~\citep{zhou2025external}:} This method introduces a dual-branch injection framework to encode external knowledge into CLIP. It utilizes data augmentation and GPT-4 descriptors to enrich visual and textual features during incremental tasks.
\item{\bf BOFA~\citep{li2025bofa}:}
This method concentrates all adaptation within CLIP’s existing cross-modal bridge-layer to avoid extra parameters and employs orthogonal low-rank fusion to accumulate knowledge without forgetting.
\end{itemize}

\end{document}